\documentclass[twoside]{article}

\DeclareUnicodeCharacter{044A}{b}   
\DeclareUnicodeCharacter{0411}{B}   
\DeclareUnicodeCharacter{043B}{l}   
\DeclareUnicodeCharacter{0433}{g}   
\DeclareUnicodeCharacter{0430}{a}   
\DeclareUnicodeCharacter{0440}{r}   
\DeclareUnicodeCharacter{0441}{s}   
\DeclareUnicodeCharacter{043A}{k}   
\DeclareUnicodeCharacter{043E}{o}   
\DeclareUnicodeCharacter{0435}{e}   
\DeclareUnicodeCharacter{0421}{S}   
\DeclareUnicodeCharacter{043F}{p}   
\DeclareUnicodeCharacter{0438}{i}   
\DeclareUnicodeCharacter{043D}{n}   
\DeclareUnicodeCharacter{0437}{z}   
\DeclareUnicodeCharacter{0410}{A}   
\DeclareUnicodeCharacter{0445}{h}   
\DeclareUnicodeCharacter{044F}{ya}  

\usepackage{amsmath}

\usepackage{PRIMEarxiv}

\usepackage[utf8]{inputenc} 
\usepackage[T1]{fontenc}    
\usepackage{hyperref}       
\usepackage{url}            
\usepackage{booktabs}       
\usepackage{amsfonts}       
\usepackage{nicefrac}       
\usepackage{microtype}      
\usepackage{lipsum}
\usepackage{fancyhdr}       
\usepackage{graphicx}       
\graphicspath{{media/}}     
\fancypagestyle{plain}{%
  \fancyhf{}%
  \fancyfoot[RO,LE]{\thepage}%
}

\title{Detecting Media Clones in Cultural Repositories Using a Positive Unlabeled Learning Approach
}

\author{\small
  \textbf{V. Sevetlidis, V. Arampatzakis, M. Karta, I. Mourthos, D. Tsiafaki, G. Pavlidis} \\
  Archimedes, ATHENA RC, Greece \\
  \texttt{\{vasiseve,vasilis.arampatzakis,melpomek,jmourthos,tsiafaki,gpavlid\}@athenarc.gr} \\
}

\begin{document}
\maketitle
\thispagestyle{plain}

\begin{abstract}
We formulate curator-in-the-loop duplicate discovery in the AtticPOT repository as a Positive–Unlabeled (PU) learning problem. Given a \emph{single anchor per artefact}, we train a lightweight per-query \emph{Clone Encoder} on augmented views of the anchor and score the unlabeled repository with an interpretable threshold on the latent $\ell_2$ norm. The system proposes candidates for curator verification, uncovering cross-record duplicates that were not verified \emph{a priori}. On CIFAR--10 we obtain $\mathrm{F1}=96.37$ ($\mathrm{AUROC}=97.97$); on AtticPOT we reach $\mathrm{F1}=90.79$ ($\mathrm{AUROC}=98.99$), improving F1 by $+7.70$ points over the best baseline (SVDD) under the same lightweight backbone. Qualitative ``find-similar'' panels show stable neighbourhoods across viewpoint and condition. The method avoids explicit negatives, offers a transparent operating point, and fits de-duplication, record linkage, and curator-in-the-loop workflows.
\end{abstract}

\keywords{Cultural heritage repositories \and Positive--Unlabeled learning \and Image de-duplication \and Visual similarity retrieval}

\section{Introduction}
Digital repositories have become central to the study of cultural heritage, offering
archaeologists, historians and the wider public unprecedented access to vast collections
of artefacts. Projects such as \emph{AtticPOT\footnote{http://AtticPOT.athenarc.gr/index.php/el/}}, which catalogues thousands of Attic
artefacts and sherds from ancient Thrace, illustrate both the promise and the challenge of
digital infrastructures: while they preserve and make accessible important cultural
assets, the sheer scale and heterogeneity of the material can overwhelm researchers.
Identifying whether two images depict the same artefact, or whether fragments belong to
a vessel type already represented in the collection, remains a largely manual and
time-consuming process.

This problem has concrete consequences. Duplicate entries across publications may
inflate statistical analyses of pottery distribution. Subtly different photographs of the
same vessel, captured under varying conditions, may be mistakenly treated as separate
items. Conversely, visually similar fragments may go unnoticed, preventing researchers
from reconstructing production patterns, workshop practices, or trade routes. Existing
digital tools in repositories such as AtticPOT already support advanced queries,
visualisation, and spatial analysis, yet they stop short of offering automated visual
similarity detection. Addressing this gap is crucial if digital heritage collections are to
evolve from passive catalogues into active research assistants.

In this paper we introduce a machine learning method tailored for this challenge,
inspired by the paradigm of \emph{Positive--Unlabeled (PU) Learning}. Unlike
conventional approaches that require extensive labelled datasets, PU learning leverages
only a handful of known positive examples (e.g., photographs of the same artefact) against
a large pool of unlabeled material. The core intuition is simple: we teach a neural
network to cluster positive examples closely together, while adaptively learning a
margin that keeps other images outside this cluster. In plain terms, the system learns to
draw a protective ``circle'' around an artefact and to flag intruders that look too similar to
ignore.

Our contributions are fourfold: (i) we formulate \textbf{artefact clone detection} in cultural-heritage repositories as a Positive--Unlabeled learning problem; (ii) we design a lightweight \textbf{Clone Encoder} that maps artefact images to compact representations and supports an adaptive margin–based decision rule on the latent $\ell_2$ norm; (iii) we show that controlled image transformations simulate realistic re-photography and preservation variation, enabling training from limited data (single-anchor setting); and (iv) we validate the approach on CIFAR--10 (mean F1 $\approx 0.96$) and demonstrate its applicability to the AtticPOT repository as a cultural-heritage case study.


The remainder of this paper is organised as follows. Section~\ref{sec:background} discusses
content based image retrieval applications within cultural heritage repositories. Section~\ref{sec:methodology}
details the learning framework and architecture. Section~\ref{sec:evaluation}~and~\ref{sec:results} report the experimental design and the results respectively. Section~\ref{sec:discussion}
reflects on limitations and future opportunities, and Section~\ref{sec:conclusion}
summarises the findings and outlines directions for future work.

\section{Background and Related Work}
\label{sec:background}
Large-scale repositories have transformed access to cultural heritage (CH) images and metadata, enabling cross-collection search and analysis. Initiatives such as \emph{Europeana} pioneered programmatic access and services like image similarity search to support exploration at scale \cite{EuropeanaIIIFKB, EuropeanaRecordAPI,EuropeanaSearchAPI, EuropeanaAPIsLanding,pavlidis2015demystifying}. The \emph{International Image Interoperability Framework} (IIIF) further established common APIs for image delivery and annotation, underpinning many GLAM (Galleries, Libraries, Archives, and Museums) platforms \cite{iiif_overview}. Within this landscape, \emph{AtticPOT} assembled a bilingual, queryable repository of Attic pottery in Thrace, with GIS and statistical tooling that facilitates distributional and contextual analyses for archaeological research \cite{AtticPOT_project,AtticPOT_rare_shapes}. While these systems provide strong foundations for discovery and mapping, most do not yet integrate learned visual similarity tailored to pottery-specific tasks (duplicate detection, stylistic grouping, fragment suggestion).

Content-based image retrieval (CBIR) has a long trajectory in CH, progressing from handcrafted descriptors to deep representations. Surveys highlight the maturation of deep features, metric learning, and scalability for retrieval \cite{dubey2020cbir,long2003fundamentals}. On the CH side, museum-centered benchmarks and services catalyzed research: the \emph{Rijksmuseum Challenge} released 110k artworks with rich metadata to stimulate recognition and retrieval tasks \cite{mensink2014rijksmuseum}; OmniArt introduced a multi-task learning setup across hundreds of thousands of museum records \cite{strezoski2017omniart}; SemArt coupled paintings with catalogue-like texts to study semantic (text–image) retrieval \cite{garcia2018semart}. Beyond art, CH platforms investigated production-grade visual search for end users \cite{europeana_similarity,klic2023linkedopenimages,arampatzakis2021art3mis}. 

Near-duplicate detection---critical for repository de-duplication and record linkage---has often relied on perceptual hashing or shallow features, with recent surveys and evaluations consolidating best practices and limitations \cite{phaser2024,perchash_survey2025,near_duplicate_survey2020}. Compared to these, learned deep embeddings can offer more robust invariances to illumination, viewpoint, or minor restoration differences, but they typically assume labeled positives and clean negatives.

A growing body of work applies deep learning to archaeological ceramics. Studies report strong performance on decorated sherd classification and fabric identification from thin sections \cite{lyons2021ceramic,chetouani2020classification}. Project-level efforts, such as ArchAIDE, integrated recognition pipelines to assist pottery identification in practice \cite{archaide2021}. Recent e-Heritage works explored relief-printed motif matching from 3D scans, combining supervised and unsupervised strategies \cite{brahim2023relief}. These contributions demonstrate feasibility but largely rely on supervised labels or curated training pairs. In contrast, repository-scale tasks (e.g., \emph{find visually similar artefact} across collections) often lack exhaustive negative labels and may contain latent duplicates.

Because expert labels are expensive, CH increasingly complements institutional datasets with web and crowdsourced contributions. Europeana and IIIF infrastructures ease aggregation; platforms like \emph{CrowdHeritage} show that community tagging/validation can measurably improve metadata quality for re-use \cite{kaldeli2021crowdheritage}. The GLAM literature documents crowdsourced transcription, tagging, and annotation campaigns (e.g., Zooniverse, BL projects) that scale curation while surfacing ethical and quality-control considerations \cite{oomen2011crowdsourcing,ridge2021chapter,ala2023crowdsourcing,blickhan2019collab,wikilovesmonuments2022}. For vision tasks, art-focused datasets (Rijksmuseum, OmniArt, SemArt, WikiArt) were often assembled from public web sources and institutional portals, then structured to support retrieval and classification benchmarks \cite{mensink2014rijksmuseum,strezoski2017omniart,garcia2018semart,wikiartvectors2022}. Another example in this area is the work by Sevetlidis et al.: they built a pipeline for curating web-acquired image datasets — specifically a Greek food image set \cite{sevetlidis2021augmenting,pavlidis2020ai} — with focus on cleaning, deduplication, and bias mitigation \cite{sevetlidis_food2021}. Through anomaly-informed and deduplication mechanisms, this pipeline enabled more reliable dataset assembly from noisy sources \cite{sevetlidis_bias2022}. Their methodology illustrates the kind of preprocessing needed before any learning system is deployed on large uncurated web datasets.

In repository contexts, strict negatives are rare: an unlabeled image could still be related to the query object. Positive–Unlabeled (PU) learning explicitly models this setting by contrasting known positives against a pool of unlabeled data, avoiding strong assumptions about negatives. Recent surveys and applications illustrate PU learning’s risk estimators \cite{kiryo2017positive} and treatment of false negatives across domains \cite{sevetlidis2024leveraging,sevetlidis2024dense}. To our knowledge, PU formulations have not been systematically evaluated for CH visual similarity and de-duplication, where they align naturally with curatorial realities (few certain positives; many uncertain candidates). Our method operationalizes this idea by learning a compact region for positives and an adaptive margin separating them from the unlabeled pool, providing a practical bridge between CH repository needs and modern representation learning.

\section{Methodology}
\label{sec:methodology}

\begin{figure}[ht]
  \centering
  \includegraphics[width=0.9\linewidth]{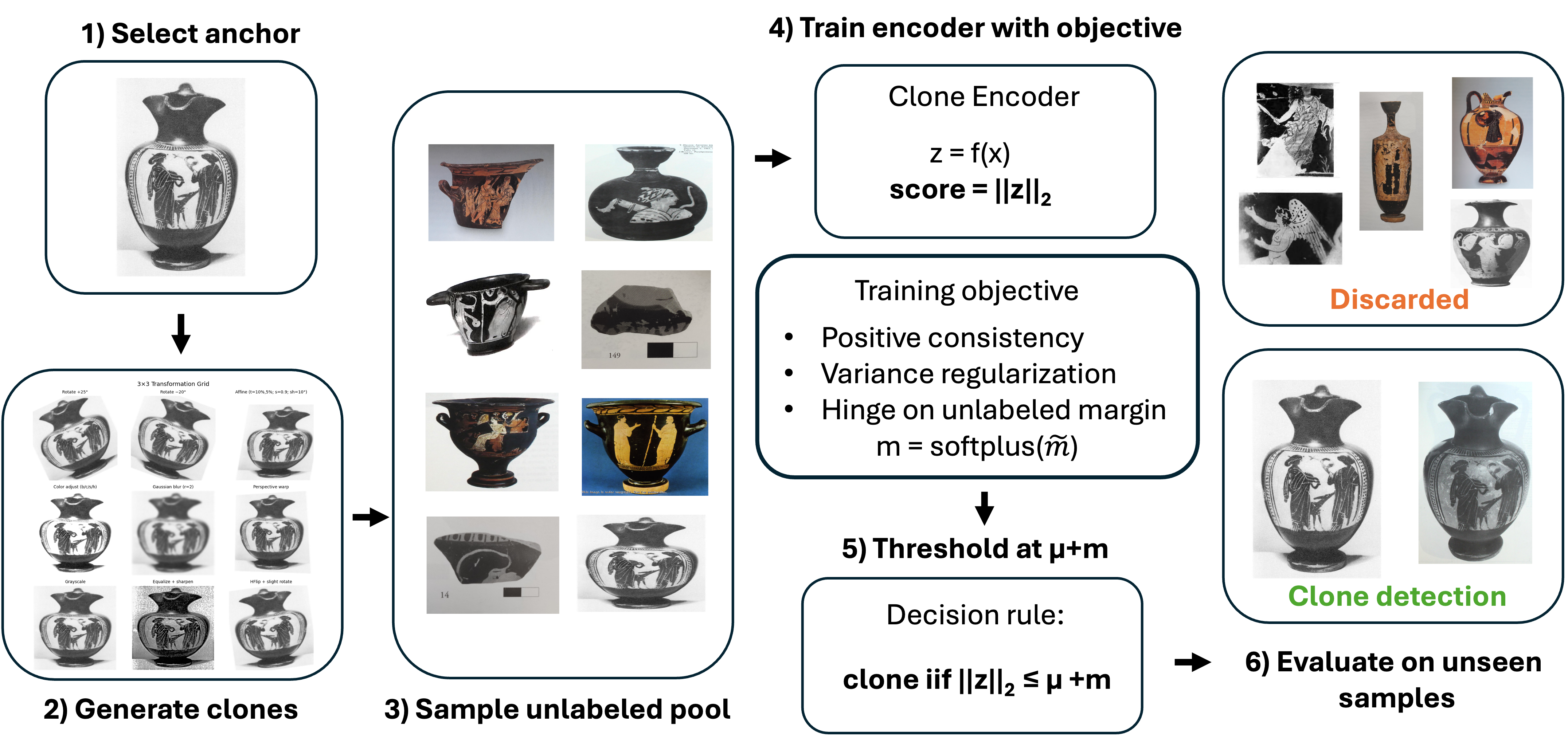} 
  \caption{Proposed clone-detection workflow: an anchor is augmented into clones and contrasted with an unlabeled pool; a Positive–Unlabeled–trained encoder discards non-matches and flags near-duplicates.}
  \label{fig:method-overview}
\end{figure}

We aim to automatically identify visually similar artefacts—``clones''—in large-scale cultural-heritage repositories such as AtticPOT, a task that has traditionally required expert, time-intensive manual work (e.g., spotting multiple occurrences of the same vessel across publications, detecting stylistic variants from the same workshop, or grouping fragments with related shapes or motifs). Our method (see, \figureautorefname~\ref{fig:method-overview}) trains a lightweight \emph{Clone Encoder} under a Positive--Unlabeled objective from a \emph{single} anchor image per artefact: augmented ``clones'' simulate re-photography and preservation variation and are contrasted against an unlabeled pool sampled from the repository. The encoder maps each image to a latent vector and operates on its $\ell_2$ norm, learning an adaptive margin that compacts positives while pushing unlabeled items beyond a data-driven threshold; an unseen image is accepted as a clone of the anchor whenever $\lVert z\rVert_2 \le \mu + m$. This simple, interpretable rule yields a practical ``find similar'' capability for repositories like AtticPOT, supporting duplicate detection across publications, grouping of stylistically related vessels, and tentative fragment attribution, thereby accelerating curatorial workflows.

\subsection{The Clone Encoder}

At the core of the approach lies a convolutional neural network (CNN) that we term the
\emph{Clone Encoder}. This network receives an image of a pottery vessel and produces
a compact numerical representation (a \emph{latent vector}) that encodes stylistic and
morphological information. The architecture follows a typical visual feature extractor:
a sequence of convolutional layers that detect edges, curves and patterns, followed by a
pooling and a linear projection step. Formally, for an input image $x$, the encoder produces a latent representation $z \in \mathbb{R}^d$. Instead of working directly with $z$, we compute its Euclidean norm $\|z\|_2 = \sqrt{\sum_{i=1}^d z_i^2}$, which condenses the information into a single scalar. This scalar will serve as the basis for deciding whether an input image is a clone of a known artefact. A key element of the architecture is a learnable \emph{margin parameter} $\tilde{m}$, transformed through the softplus function to ensure positivity: $m = \log(1 + e^{\tilde{m}})$. This margin acts as a buffer zone between known positive examples and the rest of the collection, allowing the model to adaptively set its own decision threshold.

\subsection{Positive--Unlabeled Learning}

Unlike standard supervised learning, which requires both positive and negative labels,
cultural heritage data rarely provides explicit negatives: while we may know that certain
images depict the same artefact (positives), we cannot be certain that the rest of the
repository are truly unrelated (negatives). This scenario is ideally suited to
\emph{Positive--Unlabeled (PU) Learning}, a framework that learns to separate positives
from a pool of unlabeled data.

Our learning objective is composed of three terms. Let $\mathcal{P}$ denote the set of
positives (clone images of a given artefact) and $\mathcal{U}$ the unlabeled set
(randomly sampled other images). For each batch we compute $\mu = \frac{1}{|\mathcal{P}|}\sum_{x_p \in \mathcal{P}} \| f(x_p) \|_2,
$
the mean norm of the positives. The loss function is then
\[
  \mathcal{L} =
    \underbrace{\frac{1}{|\mathcal{P}|}\sum_{x_p \in \mathcal{P}} \big(\|f(x_p)\|_2 - \mu\big)^2}_{\text{consistency term}}
    + \lambda_{\text{var}} \underbrace{\mathrm{Var}\big(\{\|f(x_p)\|_2 : x_p \in \mathcal{P}\}\big)}_{\text{variance regularization}}
    + \underbrace{\frac{1}{|\mathcal{U}|}\sum_{x_u \in \mathcal{U}} \max\{0, \mu + m - \|f(x_u)\|_2\}}_{\text{hinge loss on unlabeled}}.
\]

The \textbf{consistency term} ensures that all positive examples of the same artefact have similar norm values, effectively ``pulling'' them together. The \textbf{variance regularization} prevents excessive spread within the positive cluster. Finally, the \textbf{hinge loss} pushes the unlabeled examples outside the positive cluster, beyond the adaptive margin $\mu+m$. 

\subsection{Data Preparation and Training}

Since repositories often contain only one or a few images per artefact, we generate additional training data by applying controlled image transformations to a single photograph. These include affine transformations (rotation, scaling, shear), colour jittering, and blurring. Such augmentations simulate the variability introduced by different photographic conditions, preservation states, or digitisation processes. The training procedure alternates between presenting positive batches
(augmented clones of an artefact) and unlabeled batches (randomly sampled images from the repository, excluding the artefact itself). The Clone Encoder is trained for several epochs with stochastic gradient descent until the loss stabilises. At evaluation time,
a new image is classified as a clone if its norm satisfies   $\|f(x)\|_2 \leq \mu + m.$

\subsection{Integration with Cultural Heritage Repositories}
Applied to the AtticPOT repository, the method enables several new forms of interaction: \textbf{duplicate detection}—automatically flagging cases where the same artefact appears in multiple records or publications; \textbf{similarity search}—given an artefact image, retrieving other objects with comparable visual or stylistic features; and \textbf{fragment attribution}—suggesting potential matches between isolated sherds and known vessel types. In this way, the repository evolves from a static catalogue into an intelligent assistant for archaeological research, complementing existing GIS and statistical tools with machine-vision capabilities.

\section{Experimental Evaluation}
\label{sec:evaluation}

\subsection{Evaluation Protocol}
To assess the effectiveness of the proposed Positive--Unlabeled framework, we design an evaluation procedure that reflects the conditions of cultural heritage repositories: a small number of confirmed positive instances for a given artefact, contrasted against a large pool of unlabeled material that may contain both true negatives and unrecognized positives. Specifically, in our experiments, ``clone/duplicate'' labels are \emph{operational}, not manual: for a chosen anchor image $x$ (a single photograph of one artefact), positives are generated only by applying controlled transformations to $x$ (affine, colour jitter, blur; cf.\ \texttt{clone\_tf}); no human annotation or inter–annotator agreement is used, and we do not assert cross-publication identity. The unlabeled pool for that anchor is the set of all other usable images in the repository, i.e.\ indices $\{0,\dots,N{-}1\}\setminus\{\text{anchor}\}$ after integrity filtering; thus unlabeled may contain true negatives and occasional latent duplicates, which matches the Positive–Unlabeled setting.

\begin{figure}[ht]
  \centering
  \includegraphics[width=0.9\linewidth]{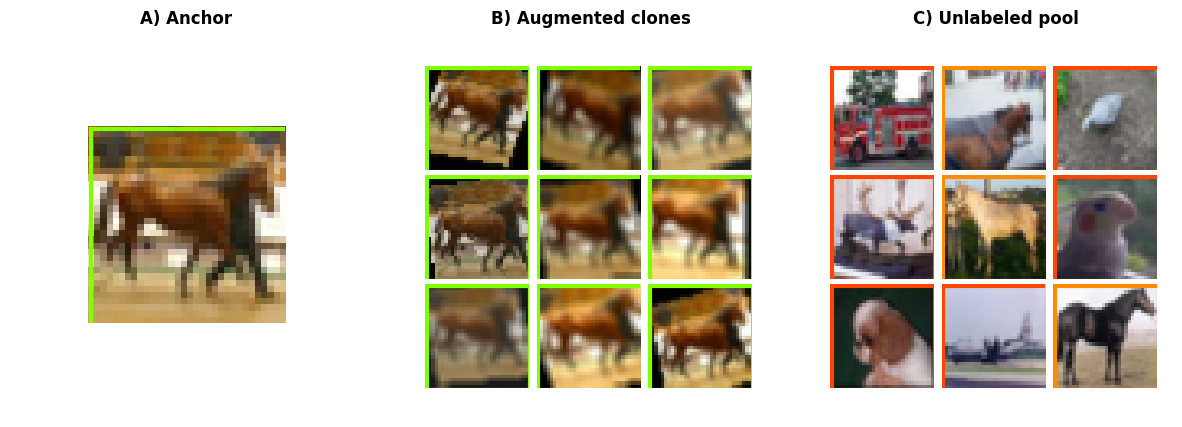}
  \small\caption{Training batch illustration on CIFAR--10. 
  Panels A: the anchor image, B: clones of the anchor via augmentation and C: unlabeled pool mixing unlabeled positives.}
  \label{fig:batch-panel}
\end{figure}

In addition to the AtticPOT dataset, which contains around 6{,}000 photographs of the project\'s documented artefacts, we adopt the CIFAR--10 image dataset, a standard benchmark widely used in the computer vision domain, as a \emph{controlled} proxy for such repositories. Although not a heritage corpus, its scale and diversity permit systematic testing of the learning dynamics under Positive--Unlabeled constraints. Each evaluation trial proceeds as follows. An \emph{anchor image} is selected, representing the artefact of interest. From this image, a set of \emph{positives} is generated by applying controlled transformations (geometric, photometric, and blur perturbations) to simulate the variability of re-photography or preservation differences. In parallel, an \emph{unlabeled set} of images is sampled from the remainder of the dataset, excluding the anchor. These unlabeled instances are not assumed to be genuine negatives; rather, they represent the uncertain background against which clone detection must operate. An example training batch is shown in Fig.~\ref{fig:batch-panel}, where the anchor (horse), its augmented clones, and a mixed unlabeled pool (including unlabeled positives and negatives) are displayed with color-coded borders.

\subsection{Training Procedure}
For each anchor image $x_a$, we instantiate and train from scratch a lightweight encoder $f_\theta:\mathbb{R}^{32\times32\times3}\!\to\!\mathbb{R}^{128}$ composed of three $5{\times}5$ convolutional layers (channels $32\!\rightarrow\!64\!\rightarrow\!128$, stride $2$, padding $2$) with \texttt{ReLU} activations, followed by adaptive average pooling to $1{\times}1$, flattening, and a linear projection to a 128-D embedding; the decision variable is the latent $\ell_2$ norm and the non-negative margin is parameterized as $m=\mathrm{softplus}(\tilde m)$. Training uses the Positive--Unlabeled objective from Section~\ref{sec:methodology}, which pulls positives toward their mean norm $\mu$ and penalizes unlabeled samples that fall within $\mu{+}m$ (variance regularization weight $\lambda_{\text{var}}{=}0$). Unless otherwise noted, inputs are $32{\times}32$ RGB images scaled to $[0,1]$ via \texttt{ToTensor} without mean–std normalization. For each anchor we form $N_p{=}128$ positive views with \texttt{RandomAffine} (rotation $\pm20^\circ$, translation $\leq10\%$ in $x/y$, scale $[0.9,1.1]$, shear $\pm10^\circ$), \texttt{ColorJitter} (brightness $0.3$, contrast $0.3$, saturation $0.2$), \texttt{GaussianBlur} (kernel size $3$), and \texttt{ToTensor}; the unlabeled set comprises $N_u{=}128$ random corpus images (anchor excluded) with only \texttt{ToTensor}. Optimization uses Adam (learning rate $10^{-3}$, no weight decay), mini-batches of $32$ positives and $32$ unlabeled per step, for $10$ epochs, with no scheduler and no early stopping; pseudo-random seeds are left at library defaults. At test time we set the threshold $\tau=\mu{+}m$ from the final training iteration and classify an unseen image $\hat x$ as a clone of $x_a$ if $\lVert f_\theta(\hat x)\rVert_2 \le \tau$. We repeat this per-anchor training over $1{,}000$ distinct anchors and report the mean precision, recall, and F1 across trials; for each anchor we probe $N_{\text{test}}{=}1000$ augmented positives and up to $N_u$ negatives drawn from the unlabeled set.

\subsection{Baseline Comparison Methods}
To situate the proposed Positive--Unlabeled framework within the broader landscape of
self-supervised representation learning, we implemented four established methods on the
same backbone architecture as our Clone Encoder:

\begin{itemize}
  \item \textbf{SimCLR} \cite{chen2020simpleframeworkcontrastivelearning}: a contrastive learning approach that maximises agreement between different augmented views of the same image while contrasting them against other images in the batch.

  \item \textbf{MoCo} (Momentum Contrast) \cite{he2020momentumcontrastunsupervisedvisual}: extends the contrastive paradigm by maintaining a dynamic memory queue of negative examples and a momentum-updated encoder to stabilise training.

  \item \textbf{BYOL} (Bootstrap Your Own Latent) \cite{grill2020bootstraplatentnewapproach}: removes the need for negative examples by encouraging consistency between an online encoder and a slowly updated target encoder.

  \item \textbf{DeepSVDD} \cite{ruff2018deep}: a deep one-class classification method that learns a representation by minimising the distance of positive instances to a centre in feature space, thereby detecting anomalies as points lying farther away.
\end{itemize}

All baselines were trained under comparable settings using the same lightweight convolutional
backbone to ensure fairness of comparison. Each was evaluated on the clone detection task
described in Section~\ref{sec:evaluation}, with classification thresholds determined by the
median similarity or score over balanced test sets.

\subsection{Testing and Metrics}
For each anchor we form two disjoint test sets: (i) augmented variants of the anchor as positives and (ii) images drawn without replacement from the unlabeled pool (anchor excluded) as putative negatives. The encoder outputs a latent norm; we define a score \(s(x)=-\lVert f_\theta(x)\rVert_2\) so that larger values are “more clone-like,” and classify an image as a clone when \(s(x)\ge -\tau\) with \(\tau=\mu+m\) from training. At this operating point we report \(\mathrm{Precision}=\frac{TP}{TP+FP}\), \(\mathrm{Recall}=\frac{TP}{TP+FN}\), and \(\mathrm{F1}=\frac{\mathrm{Precision}\cdot\mathrm{Recall}}{\mathrm{Precision}+\mathrm{Recall}}\). To characterise ranking behaviour independently of a single threshold, we sweep \(\tau\) over the score distribution and compute AUROC—the probability that a random positive receives a higher score than a random negative—and AUPRC, the area under the Precision–Recall curve, via trapezoidal integration. Unless stated otherwise we use \(N_{\text{test}}=1000\) augmented positives and \(N_{\text{test}}\) unlabeled images per anchor and average metrics over all anchors.

\section{Results}
\label{sec:results}

\begin{figure}[t]
  \centering
  \includegraphics[width=0.49\linewidth]{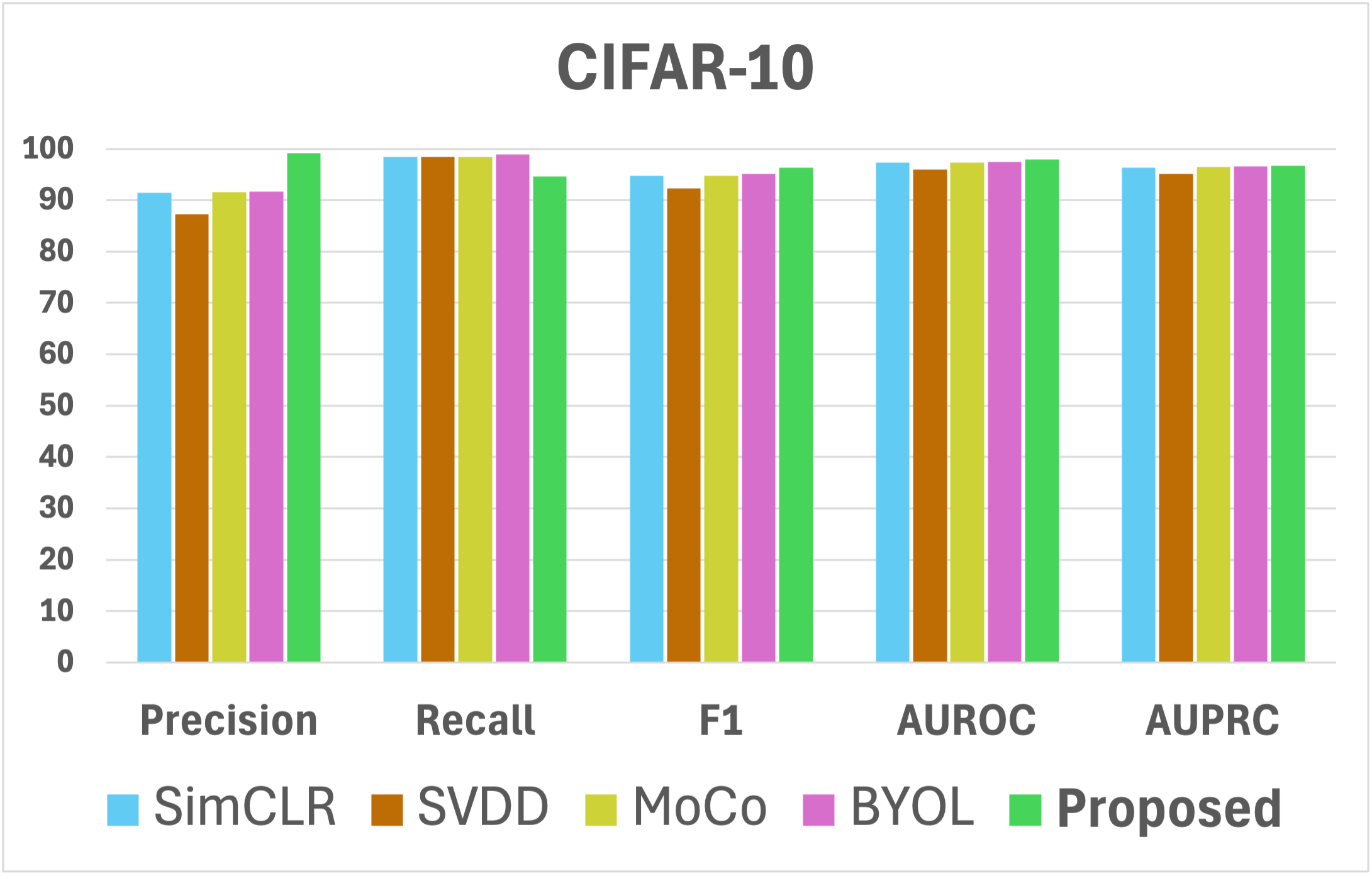}
  \includegraphics[width=0.49\linewidth]{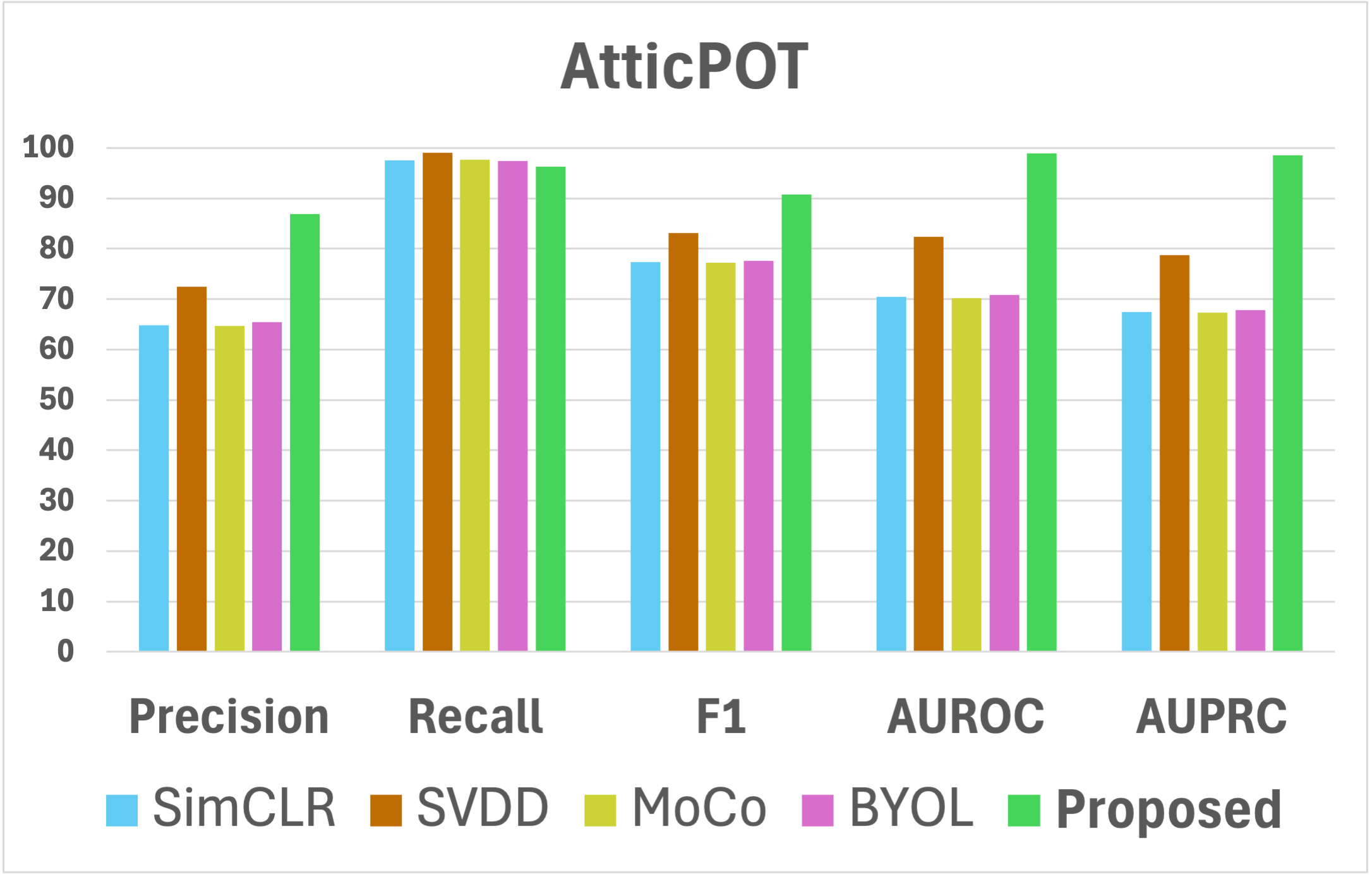}
  
  \small\caption{Duplicate detection performance on CIFAR-10 (left) and AtticPOT (right).}
  \label{fig:perf}
\end{figure}

On CIFAR--10, the \emph{Proposed} Positive--Unlabeled method achieves
$\mathrm{P}=99.19$, $\mathrm{R}=94.60$, $\mathrm{F1}=96.37$, $\mathrm{AUROC}=97.97$, and $\mathrm{AUPRC}=96.66$
(Fig.~\ref{fig:perf}, left).
Among baselines, BYOL is strongest with $\mathrm{F1}=95.09$ (SimCLR $94.71$, MoCo $94.75$, SVDD $92.32$).
Thus, \emph{Proposed} improves F1 by $+1.28$ points over the best baseline, driven primarily by a large precision gain
($+7.48$ to $+7.73$ points over SimCLR/MoCo/BYOL).
This comes with a modest recall trade--off (e.g., $-4.29$ points vs.\ BYOL), indicating a sharper operating point that reduces false positives while preserving high coverage. AUROC/AUPRC also increase (+$0.55$/+$0.15$ vs.\ BYOL), consistent with better ranking quality across thresholds.

On AtticPOT, which features heterogeneous resolutions and capture conditions, the gains are amplified
(Fig.~\ref{fig:perf}, right): \emph{Proposed} attains
$\mathrm{P}=86.89$, $\mathrm{R}=96.28$, $\mathrm{F1}=90.79$, $\mathrm{AUROC}=98.99$, and $\mathrm{AUPRC}=98.61$.
The best baseline (SVDD) reaches $\mathrm{F1}=83.09$ (SimCLR $77.34$, MoCo $77.28$, BYOL $77.65$), so \emph{Proposed}
improves F1 by $+7.70$ points (relative $+9.3\%$). The improvement concentrates in precision:
$+14.43$ points vs.\ SVDD and $+22.06$ vs.\ SimCLR, while maintaining very high recall (within $\sim\!2.7$ points of the
highest baseline recall). The ranking metrics (\mbox{AUROC $98.99$}, \mbox{AUPRC $98.61$}) far exceed the next best
(\mbox{$82.41/78.69$} for SVDD), evidencing a well-calibrated score that cleanly separates look-alikes from non-matches
across thresholds. In curatorial workflows, this translates to fewer false alerts per anchor while preserving sensitivity.

\begin{figure}[ht]
  \centering
  \includegraphics[width=0.9\linewidth]{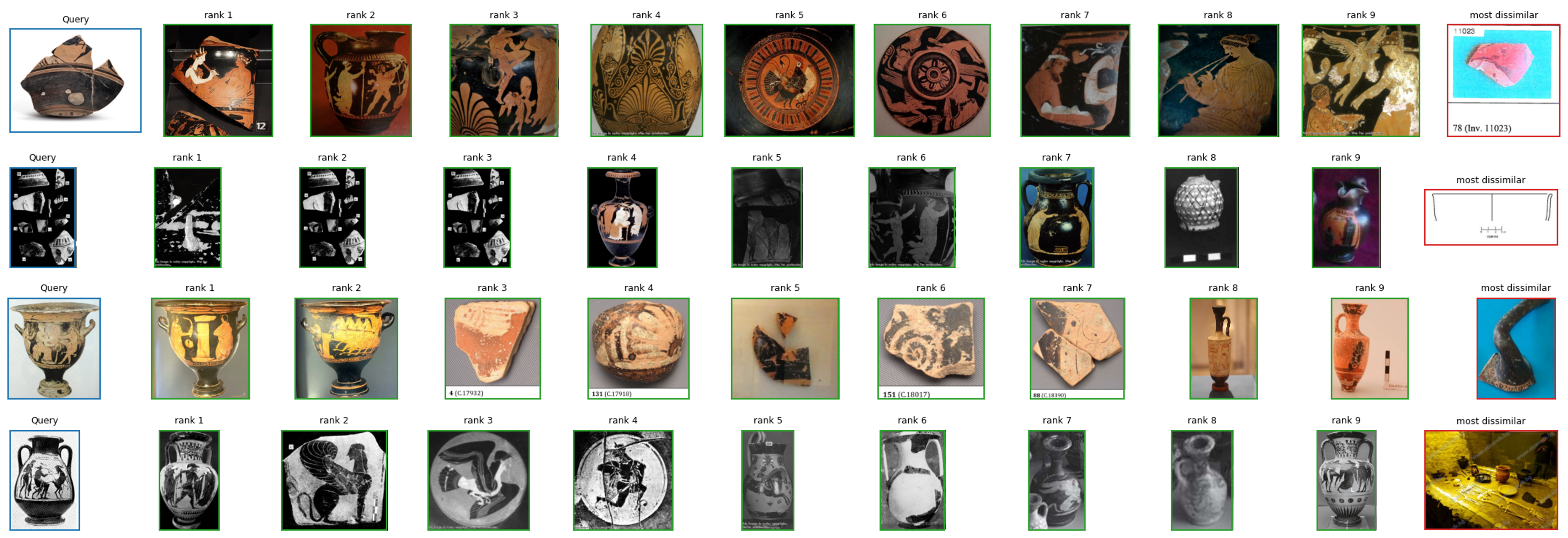}\\
  \caption{Qualitative retrieval on AtticPOT. For each query (leftmost image in a row),
  we show the nine highest-ranked images (green borders) and the single least similar (rightmost, red border).}
  \label{fig:qual-retrieval}
\end{figure}

Beyond summary metrics, we demonstrate a \emph{find–similar} use case: given a query, we train a per–query encoder with the PU objective and rank the entire repository by the learned norm (smaller $\|z\|_2$ $\Rightarrow$ more similar).
Each strip in Fig.~\ref{fig:qual-retrieval} shows the query, the top--9 retrieved images, and the most dissimilar; across diverse queries the system consistently surfaces near-duplicates and stylistically consistent views, while sending unrelated items to the tail.

\subsection{Ablation and stability studies}
\label{sec:ablations}
We evaluate four design choices: (i) variance regularisation ($\lambda_{\mathrm{var}}$), (ii) fixed vs.\ learned margin $m$, (iii) decision score (cosine-to-centroid vs.\ latent $\ell_2$), (iv) embedding dimension $d$ and weight decay. Table~\ref{tab:ablations} summarises results on the CIFAR--10 proxy (means over anchors). A \emph{fixed} small margin ($m{=}0.5$) improves the operational F1 at $\tau{=}\mu{+}m$ by $\sim$1.6 points on average compared to a learned margin, while the learned-$m$ variant reaches similar best-F1 when sweeping thresholds. Cosine-to-centroid underperforms latent $\ell_2$ across AUROC/AUPRC and best-F1. A larger embedding ($d{=}128$) slightly improves all metrics over $d{=}64$ (+0.45 F1$_\mathrm{op}$ points on average). Weight decay $10^{-4}$ and a small variance term ($\lambda_{\mathrm{var}}{=}0.1$) have negligible or slightly negative effects on F1$_\mathrm{op}$.

\begin{table}[t]
\centering
\caption{Ablations on CIFAR--10 proxy (mean over anchors).}
\label{tab:ablations}
\begin{tabular}{lcccccccc}
\toprule
Variant & $d$ & $\lambda_{\mathrm{var}}$ & $m$ & WD & F1$_\mathrm{op}$ & AUROC & AUPRC & F1$_\mathrm{best}$\\
\midrule
L2 + learned $m$               & 128 & 0.1 & learned & 0      & 0.955 & 0.998 & 0.995 & 0.991 \\
L2 + fixed $m$                 &  64 & 0.0 & 0.500   & 0      & 0.974 & 0.998 & 0.996 & 0.993 \\
L2 + fixed $m$ + WD            & 128 & 0.1 & 0.500   & 1e-04  & 0.971 & 0.999 & 0.998 & 0.994 \\
L2 + learned $m$ + $\lambda_{\mathrm{var}}$ &  64 & 0.1 & learned & 0      & 0.954 & 0.994 & 0.986 & 0.987 \\
Cosine to centroid (best-F1)   & 128 & 0.1 & 0.5     & 1e-4   & --    & 0.987 & 0.985 & 0.962 \\
\bottomrule
\end{tabular}
\end{table}

\begin{figure}[t]
  \centering
  \begin{minipage}[t]{0.32\linewidth}
    \centering
    \includegraphics[width=\linewidth]{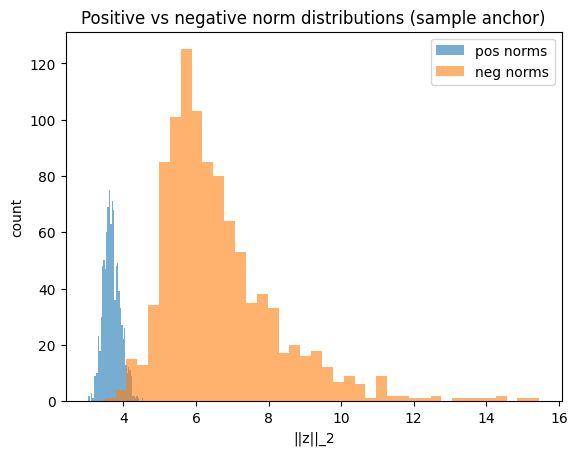}
  \end{minipage}\hfill
  \begin{minipage}[t]{0.32\linewidth}
    \centering
    \includegraphics[width=\linewidth]{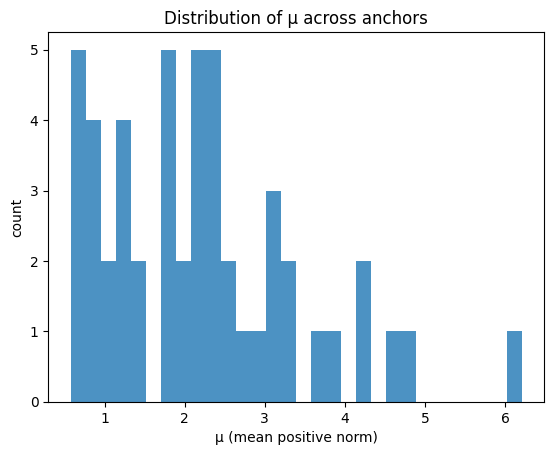}
  \end{minipage}\hfill
  \begin{minipage}[t]{0.32\linewidth}
    \centering
    \includegraphics[width=\linewidth]{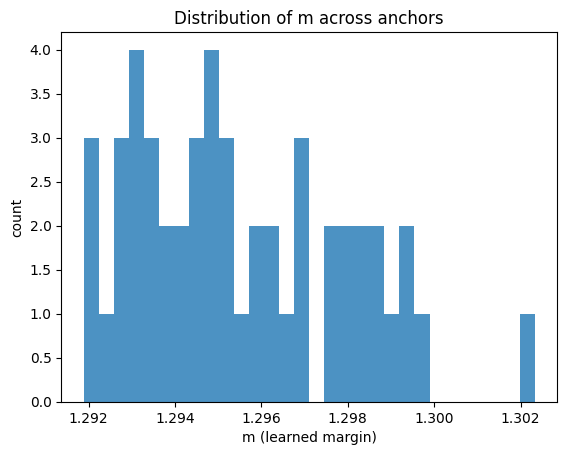}
  \end{minipage}
  \small\caption{Positive vs.\ negative latent norm distributions for a sample anchor (left). Across anchors: distribution of $\mu$ (center) and learned $m$ (right). $\mu$ varies with the anchor.}
  \label{fig:mu-m-hists}
\end{figure}

Figure~\ref{fig:mu-m-hists} (left) shows a representative anchor: positive norms concentrate in a narrow band, whereas unlabeled/negative norms are broader and shifted to larger values, yielding a clean operating gap. Across anchors, the mean positive norm $\mu$ varies (reflecting anchor-specific appearance), but the learned margin $m$ is highly stable (Fig.~\ref{fig:mu-m-hists}); for the learned-$m$, $d{=}128$ configuration we observe $\overline{m}{\approx}1.296$ with a tiny dispersion ($\sigma_m{\approx}2.5{\times}10^{-3}$ across anchors), indicating that $\mu{+}m$ is predominantly driven by the anchor-dependent $\mu$ rather than by fluctuations of $m$.

\begin{figure}[t]
  \centering
  \begin{minipage}[t]{0.32\linewidth}
    \centering
    \includegraphics[width=\linewidth]{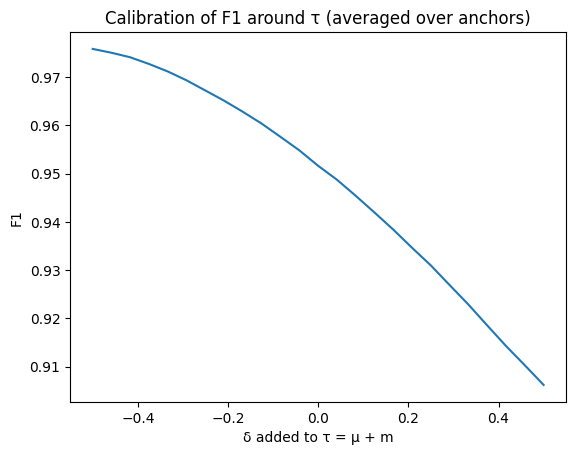}\\[-2mm]
    {\small F1 vs.\ $\delta$}
  \end{minipage}\hfill
  \begin{minipage}[t]{0.32\linewidth}
    \centering
    \includegraphics[width=\linewidth]{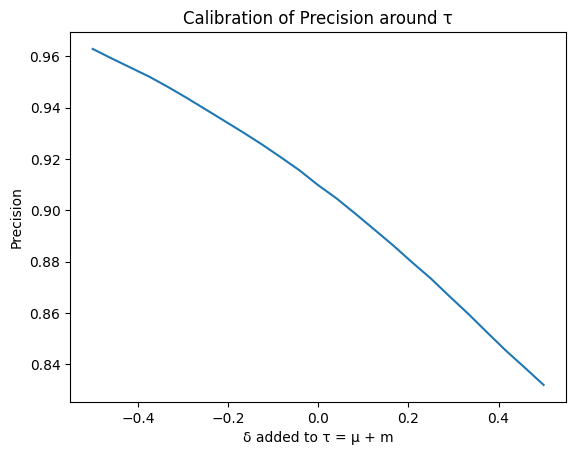}\\[-2mm]
    {\small Precision vs.\ $\delta$}
  \end{minipage}\hfill
  \begin{minipage}[t]{0.32\linewidth}
    \centering
    \includegraphics[width=\linewidth]{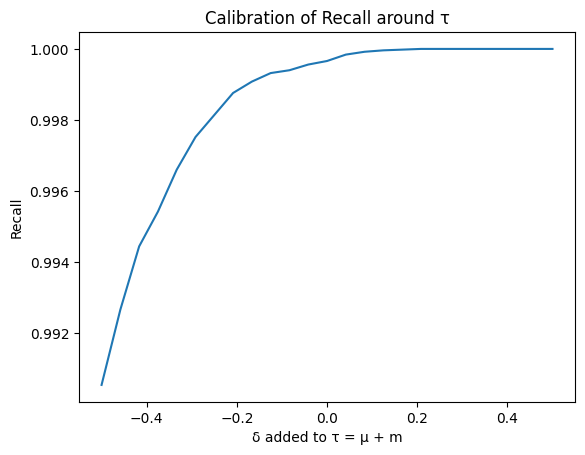}\\[-2mm]
    {\small Recall vs.\ $\delta$}
  \end{minipage}
  \caption{Calibration around the operating threshold $\tau{=}\mu{+}m$ (averaged over anchors), with $\delta\!\in\![-0.5,0.5]$.}
  \label{fig:calibration}
\end{figure}

We probe robustness by perturbing the decision threshold to $\tau{+}\delta$ and averaging metrics across anchors (Fig.~\ref{fig:calibration}). Precision decreases and recall increases monotonically as $\delta$ grows, as expected. The F1 curve exhibits a broad maximum slightly below the learned operating point (approximately $\delta\!\in\![-0.4,-0.2]$), indicating that $\tau$ is mildly permissive and that performance is stable in a neighbourhood of the learned threshold—useful when curators adjust the slider for higher precision or higher recall.

\subsection{Computing Performance}
We use a \emph{per-query} encoder: for each anchor a lightweight model is trained and the repository is scored by the latent $\ell_2$ norm (smaller $\lVert z\rVert_2$ $\Rightarrow$ more clone-like). On one CUDA GPU with $N{=}10^4$ images we measure: train $1.05$\,s (10 epochs), score $1.27$\,s, top-$k$ ($k{=}20$) $0.084$\,s; total $2.41$\,s end-to-end. This corresponds to $\approx7.9$k img/s, i.e.\ $\sim12.7$\,s for $10^5$ and $\sim127$\,s for $10^6$ images; a CPU fallback scores $10^4$ in $3.12$\,s. As our focus is clone/duplicate detection (not broad retrieval), curators typically pre-filter by publication/shape/findspot, so $N$ is often $10^4$–$5{\times}10^4$ (e.g., $50$k $\approx6.4$\,s at the measured throughput). Because encoders are query-specific, global cached features are invalid; instead we exploit batched scoring and a compact network ($\sim$0.28\,M params, $\sim$1.1\,MB) that trains in $\sim$1\,s, with latency scaling linearly in $N$ and reducible via sharding or by persisting encoders for frequent anchors. Exploring a global encoder with per-query calibration is orthogonal and left to future work; here we demonstrate that per-query fine-tuning is practical while preserving artefact-specific decision boundaries.

\section{Discussion}
\label{sec:discussion}

Cultural-heritage repositories rarely provide clean negatives: an unlabeled record may depict the same artefact under different conditions or a stylistic look-alike. Casting clone detection as Positive--Unlabeled learning matches this reality, contrasting a few confirmed positives against an uncertain background while the learned margin offers a curator-controllable boundary between a compact positive island and the broader collection. Empirically, the method improves F1 over the strongest self-supervised baseline by $+1.28$ points on CIFAR--10 and by $+7.70$ on AtticPOT (vs.\ SVDD), with gains largely driven by precision. High AUROC/AUPRC on AtticPOT (98.99/98.61) indicate that the $\ell_2$ score is well calibrated across thresholds—useful in ranking interfaces where curators browse the top of the list rather than commit to a single operating point.

Limitations include \emph{look-alike} pitfalls (distinct but similar vessels near the positive island) and \emph{anchor bias} (augmentations around a single image may under-represent artefact variability). Our per-query encoder trades compute for artefact-specific control; a global encoder with per-query calibration could reduce latency but requires additional design and is left for future work. The CIFAR--10 proxy differs from heritage imagery and AtticPOT is corpus-specific; cross-repository validation is a priority. Augmentations approximate re-photography/preservation but not extreme field conditions. For deployment, streaming/batched scoring, feature caching, and approximate nearest-neighbour indexing enable sub-second retrieval, while exposing the learned cut-off provides a single, interpretable precision–recall knob. Automated suggestions should \emph{support}—not replace—expert judgment; logging scores, thresholds, and user actions improves auditability.

Looking ahead, we aim to evolve the detector into a multimodal assistant that reasons over images and catalogue text, returning concise, sourced explanations (e.g., shape labels, fabric terms, findspots). Evaluation will combine retrieval metrics with human-centred measures (time saved, acceptance rate, explanation clarity). Governance is essential: recommendations should be traceable, avoid automatic merges, and respect provenance and cultural sensitivities; we will prototype this workflow within AtticPOT/Data-Pot to surface it in familiar repository interfaces.

\section{Conclusion}
\label{sec:conclusion}

We presented a Positive–Unlabeled framework for artefact clone detection in cultural-heritage repositories. The method trains a lightweight \emph{Clone Encoder} from a single anchor image, compacts augmented positives, and uses an adaptive margin to reject the unlabeled pool via a transparent $\ell_2$-norm decision rule. On a controlled CIFAR--10 proxy it achieved $\mathrm{F1}=96.37$ with strong ranking metrics, and on the AtticPOT repository it delivered $\mathrm{F1}=90.79$ with large precision gains over competitive self-supervised and one-class baselines, while maintaining high recall. Qualitative “find-similar’’ panels illustrate coherent neighborhoods across viewpoint and preservation variation, supporting de-duplication, record linkage, and exploratory research. The approach aligns with curatorial realities—few confirmed positives, many uncertain candidates—and exposes a single interpretable parameter for operating-point control. Looking ahead, we envision repository-integrated deployments with cached features, ANN-backed browsing, curator-in-the-loop refinement, and broader cross-collection validation. 

\section*{Acknowledgments}
This work has been partially supported by project MIS 5154714 of the National Recovery and Resilience Plan Greece 2.0 funded by the European Union under the NextGenerationEU Program.
\bibliographystyle{unsrt}  
\bibliography{references}

\end{document}